\documentclass[11pt]{article}
\usepackage{colacl}
\usepackage[dvips]{graphicx}

\title{HMM Specialization with Selective Lexicalization\thanks{The research underlying this paper was supported by research grants from Korea Science and Engineering Foundation.}}

\author{Jin-Dong Kim \and Sang-Zoo Lee \and Hae-Chang Rim \\
	Dept. of Computer Science and Engineering, Korea University, \\
	Anam-dong, Seongbuk-ku, Seoul 136-701, Korea \\
	E-mail: \{jin$\mid$zoo$\mid$rim\}@nlp.korea.ac.kr}

\begin{document}

\maketitle

\begin{abstract}
We present a technique which complements Hidden Markov Models
by incorporating some lexicalized states representing syntactically uncommon words.
Our approach examines the distribution of transitions, selects the uncommon words,
and makes lexicalized states for the words.
We performed a part-of-speech tagging experiment on the Brown corpus
to evaluate the resultant language model
and discovered that this technique improved the tagging accuracy by 0.21\%
at the 95\% level of confidence.
\end{abstract}

\section{Introduction}

Hidden Markov Models are widely used for statistical language modelling
in various fields, e.g., part-of-speech tagging or speech recognition \cite{rabiner86}.
The models are based on Markov assumptions,
which make it possible to view the language prediction as a Markov process.
In general, we make the first-order Markov assumptions
that the current tag is only dependant on the previous tag
and that the current word is only dependant on the current tag.
These are very `strong' assumptions,
so that the first-order Hidden Markov Models have the advantage of
drastically reducing the number of its parameters.
On the other hand, the assumptions restrict the model
from utilizing enough constraints provided by the local context
and the resultant model consults only a single category as the contex.

A lot of effort has been devoted in the past
to make up for the insufficient contextual information of the first-order probabilistic model.
The second order Hidden Markov Models with appropriate smoothing techniques
show better performance than the first order models
and is considered a state-of-the-art technique \cite{merialdo94,brants96}.
The complexity of the model is however relatively very high
considering the small improvement of the performance.

Garside describes IDIOMTAG \cite{garside87}
which is a component of a part-of-speech tagging system named CLAWS\@.
IDIOMTAG serves as a front-end to the tagger
and modifies some initially assigned tags
in order to reduce the amount of ambiguity to be dealt with by the tagger.
IDIOMTAG can look at any combination of words and tags,
with or without intervening words.
By using the IDIOMTAG,
CLAWS system improved tagging accuracy from 94\% to 96-97\%.
However,
the manual-intensive process of producing idiom tags is very expensive
although IDIOMTAG proved fruitful.

Kupiec \cite{kupiec92} describes a technique
of augmenting the Hidden Markov Models for part-of-speech tagging by the use of networks.
Besides the original states representing each part-of-speech,
the network contains additional states
to reduce the noun/adjective confusion,
and to extend the context for predicting past participles from preceding auxiliary verbs
when they are separated by adverbs.
By using these additional states,
the tagging system improved the accuracy from 95.7\% to 96.0\%.
However,
the additional context is chosen
by analyzing the tagging errors manually.

An automatic refining technique for Hidden Markov Models has been proposed by Brants \cite{brants96}.
It starts with some initial first order Markov Model.
Some states of the model are selected to be split or merged to take into account their predecessors.
As a result, each of new states represents a extended context.
With this technique, Brants reported a performance equivalent to the second order Hidden Markov Models.

In this paper, we present an automatic refining technique for statistical language models.
First, we examine the distribution of transitions of lexicalized categories.
Next, we break out the uncommon ones from their categories
and make new states for them.
All processes are automated
and the user has only to determine the extent of the breaking-out.

\section{``Standard'' Part-of-Speech Tagging Model based on HMM}

From the statistical point of view,
the tagging problem can be defined as
the problem of finding the proper sequence of categories $c_{1,n} = c_1 , c_2 , ... , c_n \: (n \geq 1)$
given the sequence of words $w_{1,n} = w_1 , w_2 , ... , w_n$
(We denote the {\it i}'th word by $w_i$, and the category assigned to the $w_i$ by $c_i$),
which is formally defined by the following equation:
\begin{equation}
 {\cal T}(w_{1,n}) = arg \max_{c_{1,n}} P(c_{1,n}|w_{1,n}) \label{eqn:tag}
\end{equation}

Charniak \cite{charniak93} describes the ``standard'' HMM-based tagging model as Equation \ref{eqn:hmm},
which is the simplified version of Equation \ref{eqn:tag}.
\begin{equation}
 {\cal T}(w_{1,n}) = arg \max_{c_{1,n}} \prod_{i=1}^{n} P(c_i|c_{i-1}) P(w_i|c_i) \label{eqn:hmm}
\end{equation}
With this model, we select the proper category for each word by making use of
the contextual probabilities, $P(c_i|c_{i-1})$,
and the lexical probabilities, $P(w_i|c_i)$.
This model has the advantages of a provided theoretical framework,
automatic learning facility and relatively high performance.
It is thereby at the basis of most tagging programs created over the last few years.

For this model, the first-order Markov assumtions are made as follows:
\begin{equation}
P(c_i|c_{1,i-1}, w_{1,i-1}) \approx P(c_i|c_{i-1}) \label{eqn:markov1}
\end{equation}
\begin{equation}
P(w_i|c_{1,i}, w_{1,i-1}) \approx P(w_i|c_i) \label{eqn:markov2}
\end{equation}
With Equation \ref{eqn:markov1},
we assume that the current category is independent of the previous words
and only dependent on the previous category.
With Equation \ref{eqn:markov2},
we also assume that the correct word is independent of everything except the knowledge of its category.
Through these assumptions, 
the Hidden Markov Models have the advantage of
drastically reducing the number of parameters,
thereby alleviating the sparse data problem.
However, as mentioned above, this model consults only a single category as context
and does not utilize enough constraints provided by the local context.

\section{Some Refining Techniques for HMM}

The first-order Hidden Markov Models described in the previous section provides only a single category as context.
Sometimes, this first-order context is sufficient to predict the following parts-of-speech,
but at other times (probably much more often) it is insufficient.
The goal of the work reported here is
to develop a method that can automatically refine the Hidden Markov Models
to produce a more accurate language model.
We start with the careful observation on the assumptions
which are made for the ``standard'' Hidden Markov Models.
With the Equation \ref{eqn:markov1},
we assume that the current category is only dependent on the preceding category.
As we know, it is not always true
and this first-order Markov assumption restricts the disambiguation information
within the first-order context.

The immediate ways of enriching the context are as follows:
\begin {itemize}
\item to lexicalize the context.
\item to extend the context to higher-order.
\end {itemize}
To lexicalize the context,
we include the preceding word into the context.
Contextual probabilities are then defined by $P(c_i|c_{i-1},w_{i-1})$.
\begin{figure}
\begin{center}
 \includegraphics[width=7.5cm]{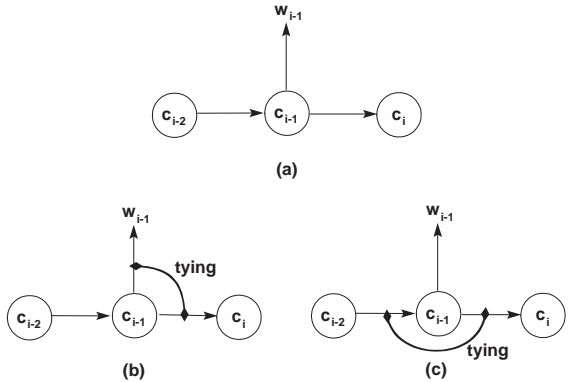}
 \caption{Two Types of Weakening the Markov Assumption}\label{fig:tying}
\end{center}
\end{figure}
Figure \ref{fig:tying} illustrates the change of dependency
when each method is applied respectively.
Figure \ref{fig:tying}(a) represents that
each first-order contextual probability and lexical probability
are independent of each other in the ``standard'' Hidden Markov Models,
where Figure \ref{fig:tying}(b) represents that
the lexical probability of the preceding word and the contextual probability of the current category
are tied into a lexicalized contextual probability.

To extend the context to higher-order,
we extend the contextual probability to the second-order.
Contextual probabilities are then defined by $P(c_i|c_{i-1},c_{i-2})$.
Figure \ref{fig:tying}(c) represents that the two adjacent contextual probabilities are tied
into the second-order contextual probability.

The simple way of enriching the context is to extend or lexicalize it uniformly.
The uniform extension of context to the second order is feasible with an appropriate smoothing technique
and is considered a state-of-the-art technique,
though its complexity is very high:
In the case of the Brown corpus, we need trigrams up to the number of 0.6 million.
An alternative to the uniform extension of context is the selective extension of context.
Brants\cite{brants96} takes this approach and reports a performance equivalent to the uniform extension
with relatively much low complexity of the model.

The uniform lexicalization of context is computationally prohibitively expensive:
In the case of the Brown corpus,
we need lexicalized bigrams up to the number of almost 3 billion.
Moreover, many of these bigrams neither contribute to the performance of the model,
nor occur frequently enough to be estimated properly.
An alternative to the uniform lexicalization is the selective lexicalization of context,
which is the main topic of this paper.

\section{Selective Lexicalization of HMM}
This section describes a new technique for refining the Hidden Markov Model,
which we call selective lexicalization.
Our approach automatically finds out syntactically uncommon words
and makes a new state (we call it a {\bf lexicalized state})for each of the words.

Given a fixed set of categories, $\{c^1 , c^2 , ... , c^C\}$,
e.g., $\{adjective, ..., verb\}$,
we assume the discrete random variable ${\bf X}_{c^j}$
with domain the set of categories
and range a set of conditional probabilities.
The random variable ${\bf X}_{c^j}$ then represents
a process of assigning
a conditional probability $P(c^i|c^j)$
to every category $c^i$ ($c^i$ ranges over ${c^1 ... c^C}$)
\begin{eqnarray}
{\bf X}_{c^j}(c^1) &= P(c^1|c^j) \nonumber \\
{\bf X}_{c^j}(c^2) &= P(c^2|c^j) \nonumber \\
&... \nonumber \\
{\bf X}_{c^j}(c^C) &= P(c^C|c^j) \nonumber
\end{eqnarray}

We convert the process of ${\bf X}_{c^j}$ into the {\bf state transition vector}, ${\bf V}_{c^j}$,
which consists of the corresponding conditional probabilities, e.g.,
\begin{center}
${\bf V}_{prep} = {\bf (} P(adjective|prep), ... , P(verb|prep) {\bf )}^T$.
\end{center}
The (squared) distance between two arbitrary vectors is then computed as follows:
\begin{equation}
R({\bf V}_1 , {\bf V}_2) = ({\bf V}_1 - {\bf V}_2)^T ({\bf V}_1 - {\bf V}_2)
\end{equation}

\begin{figure*}
\begin{center}
\input{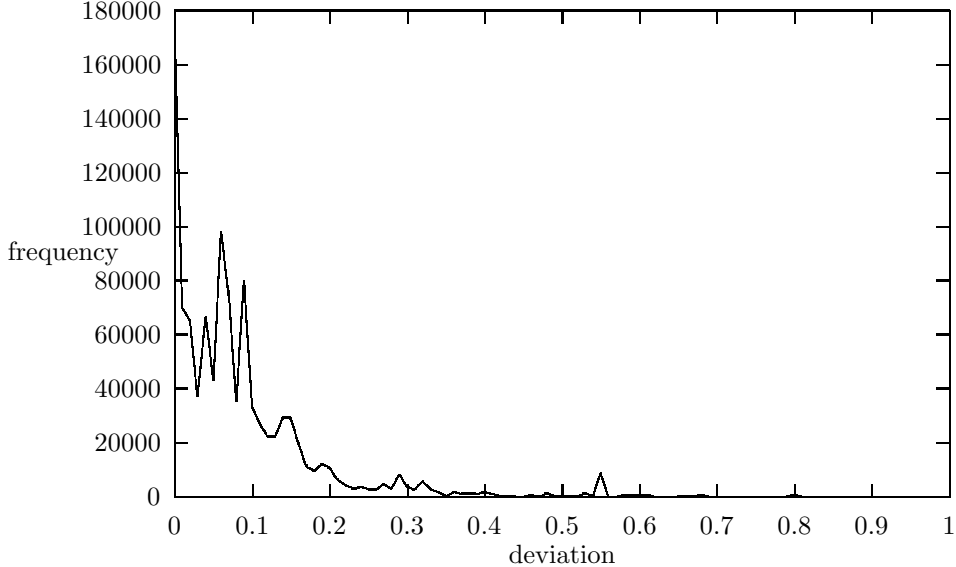}
\caption{Distribution of Lexicalized Vectors according to Deviation}\label{fig:deviation}
\end{center}
\end{figure*}

Similarly,
we define the {\bf lexicalized state transition vector}\footnote{
To alleviate the sparse data problem,
we smoothed the lexicalized state transition probabilities
by MacKay and Peto\cite{mackay95}'s smoothing technique.},
${\bf V}_{c^j,w^k}$, e.g.,
\begin{center}
${\bf V}_{prep,in}={\bf (} P(adjective|prep,in), ... , P(verb|prep,in) {\bf )}^T$.
\end{center}
In this situation,
it is possible
to regard each lexicalized state transition vector, ${\bf V}_{c^j,w^k}$, of the same category $c^j$
as members of a cluster whose centroid is the state transition vector, ${\bf V}_{c^j}$.
We can then compute the deviation of each lexicalized state transition vector, ${\bf V}_{c^j,w^k}$,
from its corresponding centroid.
\begin{equation}
D({\bf V}_{c^j,w^k}) = ({\bf V}_{c^j,w^k} - {\bf V}_{c^j})^T ({\bf V}_{c^j,w^j} - {\bf V}_{c^j})
\end{equation}
Figure \ref{fig:deviation} represents
the distribution of lexicalized state transition vectors according to their deviations.
As you can see in the figure,
the majority of the vectors are near their centroids
and only a small number of vectors are very far from their centroids.
In the first-order context model (without considering lexicalized context),
the centroids represent all the members belonging to it.
In fact, the deviation of a vector is a kind of (squared) error for the vector.
The error for a cluster is
\begin{equation}
e({\bf V}_{c^j}) = \sum_{w^k} D({\bf V}_{c^j,w^k})
\end{equation}
and the error for the overall model is simply the sum of the individual cluster errors:
\begin{equation}
E = \sum_{c^j} e({\bf V}_{c^j})
\end{equation}

Now, we could break out a few lexicalized state vectors which have large deviation ($D>\theta$)
and make them individual clusters
to reduce the error of the given model.

\begin{figure*}
\begin{center}
\input{fig-vct-IN.tex}
\input{fig-vct-IN-in.tex}
\input{fig-vct-IN-with.tex}
\input{fig-vct-IN-out.tex}
\caption{Transition Vectors in {\bf preposition} Cluster}\label{fig:vct-ex}
\end{center}
\end{figure*}
As an example, let's consider the {\bf preposition} cluster.
The value of each component of the centroid, ${\bf V}_{prep}$, is illustrated in Figure \ref{fig:vct-ex}(a)
and that of the lexicalized vectors,
${\bf V}_{prep,in}$, ${\bf V}_{prep,with}$ and ${\bf V}_{prep,out}$
are in Figure \ref{fig:vct-ex}(b), (c) and (d)
respectively.
As you can see in these figures,
most of the prepositions including {\em in} and {\em with}
are immediately followed by article(AT), noun(NN) or pronoun(NP),
but the word {\em out} as preposition shows a completely different distribution.
Therefore, it would be a good choice to break out the lexicalized vector, ${\bf V}_{prep,out}$,
from its centroid, ${\bf V}_{prep}$.

\begin{figure}
\begin{center}
 \includegraphics[width=7.5cm]{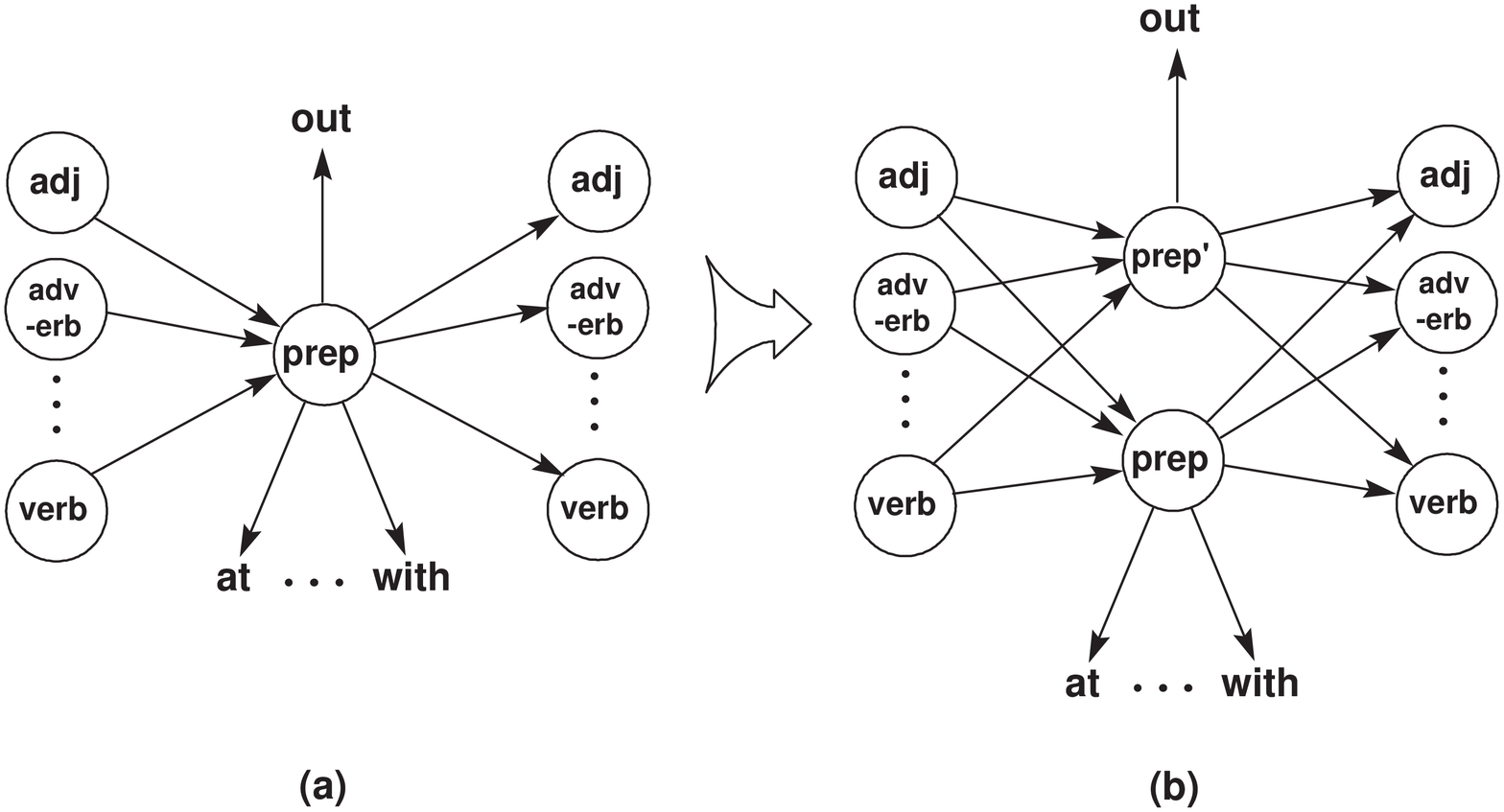}
 \caption{Splitting the {\bf preposition} State}\label{fig:split}
\end{center}
\end{figure}

From the viewpoint of a network, the state representing {\bf preposition} is split into two states;
the one is the state representing ordinary prepositions except {\em out},
and the other is the state representing the special preposition {\em out},
which we call a {\bf lexicalized state}.
This process of splitting is illustrated in Figure \ref{fig:split}.

Splitting a state results in some changes of the parameters.
The changes of the parameters resulting from lexicalizing a word, $w^k$, in a category, $c^j$,
are indicated in Table \ref{tbl:fsplit}
($c^i$ ranges over ${c^1 ... c^C}$).
\begin{table}[h]
 \caption{Changes of Parameters in Full Splitting}\label{tbl:fsplit}
\begin{center}
\begin{tabular}{|c|c|} \hline
before splitting & after splitting        \\ \hline \hline
$P(w^i|c^j)$   & $P(w^i | c^j, w^k)$      \\
               & $P(w^i | c^j, \neg w^k)$ \\ \hline
$P(c^i|c^j)$   & $P(c^i | c^j, w^k)$      \\
               & $P(c^i | c^j, \neg w^k)$ \\ \hline
$P(c^j|c^i)$   & $P(c^j, w^k | c^i)$      \\
               & $P(c^j, \neg w^k | c^i)$ \\ \hline
\end{tabular}
\end{center}
\end{table}
This full splitting will increase the complexity of the model rapidly,
so that estimating the parameters may suffer from the sparseness of the data.

To alleviate it, we use the pseudo splitting which leads to relatively small increment of the parameters.
The changes of the parameters in pseudo splitting are indicated in Table \ref{tbl:psplit}.
\begin{table}[h]
 \caption{Changes of Parameters in Pseudo Splitting}\label{tbl:psplit}
\begin{center}
\begin{tabular}{|c|c|} \hline
before splitting & after splitting        \\ \hline \hline
$P(w^i|c^j)$   & $P(w^i | c^j)$           \\ \hline
$P(c^i|c^j)$   & $P(c^i | c^j, w^k)$      \\
               & $P(c^i | c^j, \neg w^k)$ \\ \hline
$P(c^j|c^i)$   & $P(c^j | c^i)$           \\ \hline
\end{tabular}
\end{center}
\end{table}

\section{Experimental Result}

We have tested our technique through part-of-speech tagging experiments
with the Hidden Markov Models which are variously lexicalized.
In order to conduct the tagging experiments,
we divided the whole Brown (tagged) corpus containing 53,887 sentences (1,113,191 words) into two parts.
For the {\bf training set}, 90\% of the sentences were chosen at random,
from which we collected all of the statistical data.
We reserved the other 10\% for testing.
Table \ref{tbl:corpus} lists the basic statistics of our corpus.
\begin{table}[h]
 \caption{Overview of Our Corpora}\label{tbl:corpus}
\begin{center}
\begin{tabular}{|c||c|c|} \hline
               & \# of sentences & \# of words \\ \hline \hline
 training set  &     48,499      &  1,001,712  \\ \hline
  test set     &      5,388      &    111,479  \\ \hline
\end{tabular}
\end{center}
\end{table}

We used a tag set containing 85 categories.
The amount of ambiguity of the test set is summarized in Table \ref{tbl:amb}.
The second column shows that words to the ratio of 52\% (the number of 57,808) are not ambiguous.
The tagger attempts to resolve the ambiguity of the remaining words.
\begin{table}[h]
 \caption{Amount of Ambiguity of Test Set}\label{tbl:amb}
\begin{center}
\begin{tabular}{|c||c|c|c|c|c|c|c|} \hline
 ambiguity(\#)  & 1  & 2  & 3 & 4 & 5 & 6 & total \\ \hline
  ratio(\%)     & 52 & 30 & 8 & 7 & 2 & 1 & 100   \\ \hline
\end{tabular}
\end{center}
\end{table}

\begin{figure*}
\begin{center}
 \includegraphics[height=5.5cm]{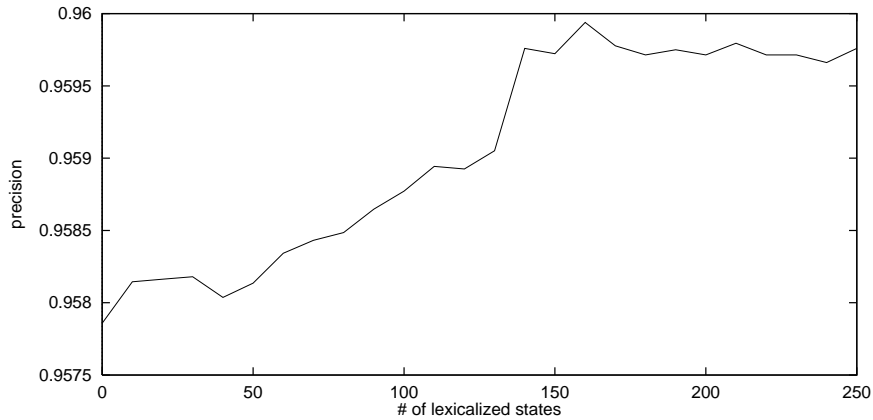}
 \caption{POS tagging results with lexicalized HMM using full splitting method}\label{fig:result1}
\end{center}
\end{figure*}

\begin{figure*}
\begin{center}
 \includegraphics[height=5.5cm]{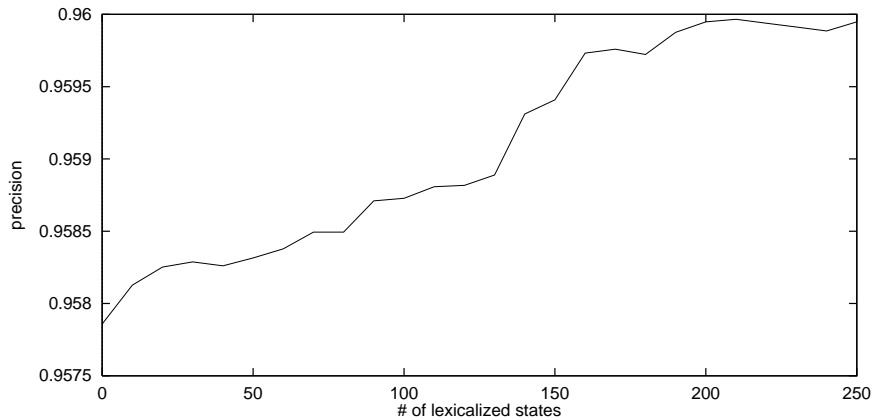}
 \caption{POS tagging results with lexicalized HMM using pseudo splitting method}\label{fig:result2}
\end{center}
\end{figure*}

Figure \ref{fig:result1} and Figure \ref{fig:result2} show
the results of our part-of-speech tagging experiments
with the ``standard'' Hidden Markov Model and variously lexicalized Hidden Markov Models
using full splitting method and pseudo splitting method respectively.

We got 95.7858\% of the tags correct
when we applied the standard Hidden Markov Model without any lexicalized states.
As the number of lexicalized states increases, the tagging accuracy increases
until the number of lexicalized states becomes 160 (using full splitting)
and 210 (using pseudo splitting).
As you can see in these figures, the full splitting improves the performance of the model more rapidly
but suffer more sevelery from the sparseness of the training data.
In this experiment,
we employed Mackay and Peto's smoothing techniques for estimating the parameters required for the models.
The best precision has been found to be 95.9966\% through the model with the 210 lexcalized states
using the pseudo splitting method.

\section{Conclusion}

In this paper, we present a method for complementing the Hidden Markov Models.
With this method, we lexicalize the Hidden Markov Model seletively and automatically
by examining the transition distribution of each state relating to certain words.

Experimental results showed that the selective lexicalization improved the tagging accurary
from about 95.79\% to about 96.00\%.
Using normal tests for statistical significance
we found that the improvement is significant at the 95\% level of confidence.

The cost for this improvement is minimal.
The resulting network contains 210 additional lexicalized states
which are found automatically.
Moreover, the lexicalization will not decrease the tagging speed\footnote{
The Viterbi algorithm for finding the best tags runs in $O(n^2)$
where $n$ is the number of states.},
because the lexicalized states and their corresponding original states are exclusive in our lexicalized network,
and thus the rate of ambiguity is not increased even if the lexicalized states are included.

Our approach leaves much room for improvement.
We have so far considered only the outgoing transitions from the target states.
As a result, we have discriminated only the words with right-associativity.
We could also discriminate the words with left-associativity
by examining the incoming transitions to the state.
Furthermore, we could extend the context by using the second-order context
as represented in Figure \ref{fig:tying}(c).
We believe that the same technique presented in this paper could be applied to the proposed extensions.

\bibliographystyle{acl}
\bibliography{mylib}
\nocite{church88}
\nocite{derose88}
\nocite{kupiec92}

\section*{\\ Appendix :\\ \center{Top 100 words with high deviation}}
\input{lexword.lst}

\end{document}